# A GABOR FILTER TEXTURE ANALYSIS APPROACH FOR HISTOPATHOLOGICAL BRAIN TUMOUR SUBTYPE DISCRIMINATION

Omar Sultan Al-Kadi

King Abdullah II School for Information Technology, University of Jordan
Amman 11942 Jordan
(o.alkadi@ju.edu.jo)

**Abstract**

Meningioma brain tumour discrimination is challenging as many histological patterns are mixed between the different subtypes. In clinical practice, dominant patterns are investigated for signs of specific meningioma pathology; however the simple observation could result in inter- and intra-observer variation due to the complexity of the histopathological patterns. Also employing a computerised feature extraction approach applied at a single resolution scale might not suffice in accurately delineating the mixture of histopathological patterns. In this work we propose a novel multiresolution feature extraction approach for characterising the textural properties of the different pathological patterns (i.e. mainly cell nuclei shape, orientation and spatial arrangement within the cytoplasm). The patterns' textural properties are characterised at various scales and orientations for an improved separability between the different extracted features. The Gabor filter energy output of each magnitude response was combined with four other fixed-resolution texture signatures (2 model-based and 2 statistical-based) with and without cell nuclei segmentation. The highest classification accuracy of 95% was reported when combining the Gabor filters' energy and the meningioma subimage fractal signature as a feature vector without performing any prior cell nuceli segmentation. This indicates that characterising the cell-nuclei self-similarity properties via Gabor filters can assists in achieving an improved meningioma subtype classification, which can assist in overcoming variations in reported diagnosis.

# 1. INTRODUCTION

Medical image analysis has contributed significantly to anatomical pathology, reveling novel pathways in histopathological examination of samples for accurate diagnosis of cancer and other related diseases. Analyzing the texture patterns is a useful way to extract meaningful features that represent underlying pathology. The main concern in texture analysis is to capture distinctive features that can maximize the difference between the analysed images and subsequently facilitate the pattern classifiers' mission. Each feature extraction method has its own unique trend for detecting discontinuities in image texture, yet its efficiency is determined by how it formulates the relationship between primary image elements, i.e., textons.

Characterising texture from a multiresolution perspective has the advantage of filtering-out noise while simultaneously giving more emphasis to features that contribute better to subtype distinction. Wavelets and filter banks such as Gabor filters are examples of multiresolution techniques that can break down texture statistical complexity [1]. Also their high sensitivity to local features facilitates the processes of preattentive or subtle texture discrimination as well [2]. Moreover, according to the uncertainty principle, the wavelet transform and Gabor function can simultaneously maintain a good boundary accuracy and frequency response [3].

Gabor filters are designed to resemble the performance of the mammalian visual cortical cells, in a sense of extracting features at different orientations and scales. This multiresolution sensitivity of Gabor filters may be helpful for extracting useful meaningful features that can characterise underlying physiology. Features are derived from Gabor coefficients, as they cannot be used explicitly for texture analysis due to their high variability. Texture signatures such as entropy or local energy [4, 5], histograms and second order statistic derived from co-occurrence matrix [6, 7] were mostly used to characterise the different spatial-frequency decompositions to provide a better spatial-frequency localisation. In this work the fractal dimension (FD), which has proved to be useful in histopathological image analysis [1, 8, 9], is used as a supporting feature (signature) to complement the texture characterisation capability of the Gabor filter energy output. Also, in order to benchmark the performance of the proposed multiresolution technique, the output energy of the set of Gabor filter banks was also combined with other statistical and model based techniques, and used for histopathological brain tumours subtype classification.

## 2. METHODOLOGY

*2.1 Histopathological image acquisition*

Four subtypes of grade I meningioma tissue biopsies distinguished according to the World Health Organisation grading system [10] are used in this work, see Fig. 1. Each subtype has its own textural features (explained in Table I) that pathologists look for in the processes of tumour classification [11]. The diagnostic tumour samples were derived from neurosurgical resections at the Bethel Department of Neurosurgery, Bielefeld, Germany for therapeutic purposes. Four micrometer thick microtome sections were dewaxed on glass slides, stained with Mayer's haemalaun and eosin (H&E), dehydrated and cover-slipped with mounting medium (Eukitt®, O. Kindler GmbH, Freiburg, Germany). Archive material of cases from the years 2004 and 2005 were selected to represent typical features of each meningioma subtype. Slides were analysed on a Zeiss Axioskop 2 plus microscope with a Zeiss Achroplan 40/0.65 oil immersion objective. After manual focusing and automated background correction, images were taken at a resolution of $1300 \times 1030$ pixels, 24 bits, true colour RGB at standardised 3200 K light temperature in TIF format using Zeiss AxioVision 3.1 software and a Zeiss AxioCam HRc digital colour camera (Carl Zeiss AG, Oberkochen, Germany). Five typical cases were selected for each diagnostic group and four different photomicrographs were taken of each case, resulting in a set of 80 pictures. Each original picture was truncated to $1024 \times 1024$ pixels and then subdivided in a $2 \times 2$ subset of $512 \times 512$ pixel pictures. This resulted in a database of 320 sub-images for further analysis.

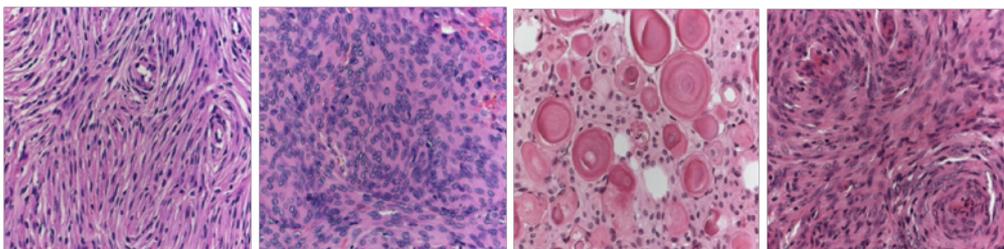

**Fig. 1. (left-right) Meningioma fibroblastic, meningothelial, psammomatous and transitional subtypes.**

TABLE I MAIN HISTOLOGICAL TEXTURAL FEATURES FOR MENINGIOMA SUBTYPES

| Subtype | Characteristics |
|---|---|
| Fibroblastic ($M_F$) | Spindle-shaped cells resembling fibroblasts in appearance, with abundant amounts of pericellular collagen. |
| Meningothelial ($M_M$) | Broad sheets or lobules of fairly uniform cells with round or oval nuclei. |
| Psammomatous ($M_P$) | A variant of transitional meningiomas with abundant psammoma bodies and many cystic spaces. |
| Transitional ($M_T$) | Contains whorls, few psammoma bodies and cells having some fibroblastic features (i.e. spindle-shaped cells) |

The relation between the fractal features and the underlying biological features manifest the discrimination between the four subtypes. Distinctive characteristics such as the small whirlpools in the form of spinning body of nuclei related to the transitional and appearing subtle in the fibroblastic subtypes; the round collection of calcium, known as psammoma bodies, that occurs repeatedly and in varying sizes in the psammomatous subtype; and the enlarged oval shape nuclei in the meningothelial which are condensed in approximate clusters, all present potential fractal characteristics that can be exploited and combined with the Gabor filter energy output.

*2.2 Gabor filter for texture analysis*

Image texture can also be analysed in a multiresolution representation using Gabor filters. It can be defined as a Gaussian modulated sinusoid with a capability of multiresolution decomposition due to its localisation both in spatial and spatial-frequency domain. Making use of Denis Gabor's class of harmonic oscillating functions within Gaussian envelopes [12], Daugman showed that the orientation and spatial-frequency selective receptive field properties of neurons in the brain's primary visual cortex can be simply modeled by 2-D Gabor-like filters [13, 14]. Bovik et al proposed a model for locating filters by exploiting individual texture power spectrum characteristics [15], while Jain and Farrokhnia further proposed a dyadic Gabor filter bank covering the spatial-frequency domain with multiple orientations [16]. Other studies proved Gabor filter to be very useful in detecting texture frequency and orientation as well [2, 17, 18] and references cited therein. The real impulse response of a 2-D sinusoidal plane wave with orientation $\theta$ and radial centre frequency $f_o$ modulated by a Gaussian envelope with standard deviations $\sigma_x$ and $\sigma_y$ respectively along the *x* and *y* axes is given by

$$h(x,y) = \frac{1}{2\pi\sigma_x\sigma_y} exp\left\{-\frac{1}{2}\left[\frac{x^2}{\sigma_x^2} + \frac{y^2}{\sigma_y^2}\right]\right\} \cos(2\pi f_o x) \qquad (1)$$

$where\ x = xcos\theta + ysin\theta$
$\qquad\quad y = -xsin\theta + ycos\theta$

The Gabor filter in the corresponding spatial-frequency domain would be represented as two symmetrically spaced Gaussians as follows

$$H(u,v) = exp\{-2\pi^2[(u-f_0)^2\sigma_x^2 + v^2\sigma_y^2]\} + exp\{-2\pi^2[(u+f_0)^2\sigma_x^2 + v^2\sigma_y^2]\} \qquad (2)$$

and spatial and corresponding spatial-frequency response are graphically shown bellow in Fig. 2.

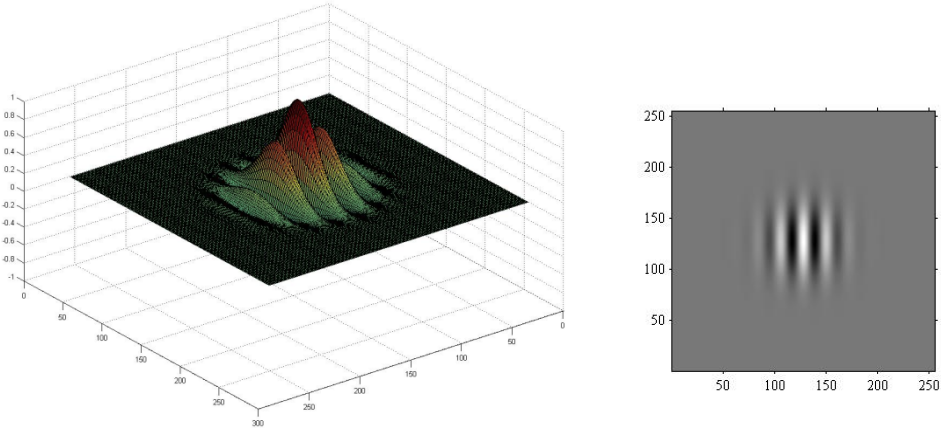

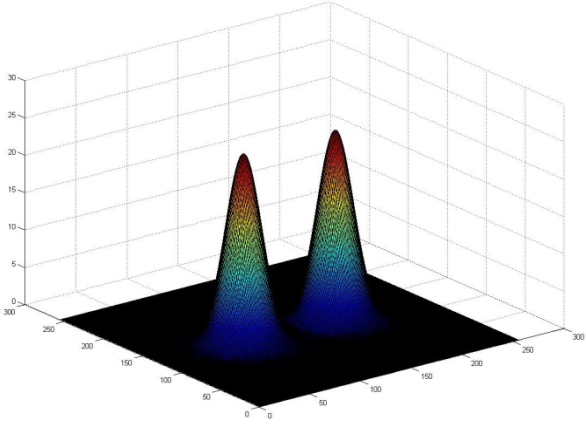

**Fig. 2 Gabor filter having 0° orientation and 32 cycles / image width for a 256 × 256 size image.**

Multiple filters covering the spatial-frequency domain can be generated by varying the filters' centre location through tuning the frequency $f_0$ with a specific angle $\theta$. Fig. 3 shows the frequency response of the dyadic filter bank in the spatial-frequency domain. The Gaussian envelope unknowns $\sigma_x$ and $\sigma_y$ can be determined as in (3) after setting frequency cut-off to -6 db and the frequency and orientation bandwidths $(B_f, B_\theta)$ to constant values matching psychovisual data [1]. In particular an interval of one octave between the radial frequencies is recommended [16]. The frequency bandwidth specified by octaves (i.e. the interval between $f_1$ and its double $f_2$) increases in a logarithmic fashion given by $\log_2(f_2/f_1)$. This was inspired by experiments that showed the frequency bandwidth of simple cells in the visual cortex is roughly one octave [19]. For this work a circular Gaussian was chosen by setting $\sigma_x = \sigma_y$ to have an equal spatial coverage in all directions, and a 45° orientation bandwidth.

$$\sigma_x = \frac{\sqrt{ln2}(2^{B_f}+1)}{\sqrt{2}\pi f_0(2^{B_f}-1)} \qquad \sigma_y = \frac{\sqrt{ln2}}{\sqrt{2}\pi f_0 tan(B_\theta/2)} \qquad (3)$$

Carefully setting the filter characteristics would result in proper capture of texture information and reduce the effect of aliasing. This is achieved by correctly selecting the filter position $(f_0, \theta)$ and bandwidth $(\sigma_x, \sigma_y)$, and making sure the central frequencies of channel filters lie close to characteristic texture frequencies to prevent the filter response from falling off too rapidly [20]. From each of the images having size of 512 × 512 used in this work, the mean was first subtracted to reduce the filter's sensitivity to texture with constant variation, then six radial frequencies ( $2^2\sqrt{2}$, $2^3\sqrt{2}$, $2^4\sqrt{2}$, $2^5\sqrt{2}$, $2^6\sqrt{2}$ $and$ $2^7\sqrt{2}$ cycles/image-width) with four orientations ($0°, 45°, 90°$ $and$ $135°$) was adopted according to [16], giving a total of 24 filters. In general, the number of dyadic Gabor filter banks required is given by $A \times \log_2(N_c/2)$, where $N_c$ is the image width and $A$ is the number of orientation separations (e.g. $A = 4$ for a 45° orientation separation angle). Filters with radial frequencies $1\sqrt{2}$ and $2\sqrt{2}$ where excluded due to their insensitivity (i.e. the filters capture spatial variations that are too large to explain textural variation in an image). Also the highest frequency was selected to be $(N_c/4)\sqrt{2}$ in order to guarantee that the passband of the filter falls inside the image. Finally the extracted features would represent the energy of each magnitude response.

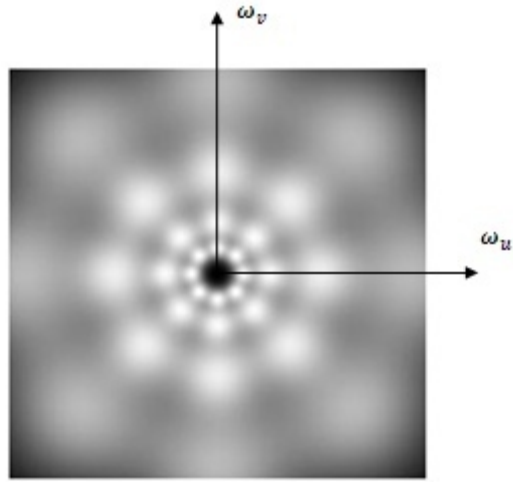

**Fig. 3 Gabor filter defined in the spatial-frequency domain with radial frequencies 1 octave apart, bandwidth 1 octave each and 45° orientation separation (only filters' half-peak supports are displayed).**

Applying a set of filter banks resembles the operation of wavelet transforming an image at selected spatial frequencies. In a way, the Gaussian function is modulated and translated for generation of the Gabor basis functions, in analogy to the scaling and translation of the mother wavelet and scaling function for wavelet basis generation. However, the Gabor function is considered an admissible wavelet [16], namely the basis produced by the Gabor function is non-orthogonal resulting in redundant decompositions. Also, depending on the size of the processed image, the number of required radial frequencies for positioning the centres of the Gabor filter banks needs to be specified prior to processing, which is similar to choosing the number of decomposition levels for the wavelet packets.

The classical method for extraction of Gabor filter texture signature is the energy $E_k$, $k$ =1, 2 in the form of $l_1$-norm [21] and $l_2$-norm [22] as in (6.12), where $M$ and $N$ are the size of the subband intensity $I_f(x, y)$.

$$E_k = \frac{1}{MN} \sum_{y=0}^{M-1} \sum_{x=0}^{N-1} |I_f(x,y)|^k \qquad (4)$$

## 3. RESULTS

The energy signature of each filter output was computed and used as a feature vector for classification. Also, the Gabor filter energy extracted features were combined with four other texture measures of the processed image itself, and the fused signatures were used for classification using a Bayesian classifier and cross-validated by leave-one-out approach. Model based signatures such as the fractal dimension (FD) and Gaussian Markov random fields (GMRF), and statistical signatures as the co-occurrence matrix (CM) and the run length matrix (RLM) were used in the combinations. As a mean for comparison, this procedure was initially applied on the blue colour channel of each image – as it showed better performance in terms of classification accuracy due to the dyes used in staining the meningioma biopsies [23] – and then applied again on the same colour channel after having the cell nuclei general structure segmented.

Table II presents the results, where the subscript S and NS indicates whether if the cell nuclei were *a priori* segmented (S) as in [23] or no segmentation was involved (NS) before extraction of the Gabor filter energy signature $G_f(E)$ alone or in combination with other texture measures. The highest classification accuracy of 95.00% was achieved when combining the image Gabor filters' energy and the fractal characteristics $G_f(E \& FD)_{NS}$ as a feature vector without performing any prior segmentation.

TABLE II CLASSIFICATION ACCURACY COMPARISONS OF GABOR FILTERED MENINGIOMA IMAGES ($G_f$) AFTER EXTRACTION OF ENERGY (E) AND IN COMBINATION WITH FRACTAL DIMENSION (FD), GAUSSIAN MARKOV RANDOM FIELDS (GMRF), CO-OCCURRENCE (CM) AND RUN-LENGTH (RLM) MATRICES TEXTURE MEASURES

| Filter texture signature | Meningioma subtype | | | | Total Accuracy |
|---|---|---|---|---|---|
| | $M_F$ | $M_M$ | $M_P$ | $M_T$ | |
| $G_f(E)_{S*}$ | 93.75 | 82.50 | 96.25 | 85.00 | 89.38% |
| $G_f(E\ \&\ FD)_S$ | 91.25 | 82.50 | 95.00 | 83.75 | 88.12% |
| $G_f(E\ \&\ GMRF)_S$ | 95.00 | 81.25 | 95.00 | 90.00 | 90.31% |
| $G_f(E\ \&\ CM)_S$ | 93.75 | 78.75 | 95.00 | 87.50 | 88.75% |
| $G_f(E\ \&\ RLM)_S$ | 92.50 | 78.75 | 98.75 | 90.00 | 90.00% |
| $G_f(E)_{NS}$ | 83.75 | 86.25 | 90.00 | 73.75 | 83.44% |
| **$G_f(E\ \&\ FD)_{NS}$** | **100** | **87.50** | **96.25** | **96.25** | **95.00%** |
| $G_f(E\ \&\ GMRF)_{NS}$ | 97.50 | 77.50 | 97.50 | 92.50 | 91.25% |
| $G_f(E\ \&\ CM)_{NS}$ | 100 | 92.50 | 96.25 | 86.25 | 93.75% |
| $G_f(E\ \&\ RLM)_{NS}$ | 100 | 80.00 | 97.50 | 82.50 | 90.00% |

\* S stands for pre-segmented cell nuclei from the background cytoplasm as in the work of [23], and NS for non-segmented cell nuclei (i.e. using the whole texture image).

The best approach – highlighted in bold in Table II – were further compared with other methods that used the same meningioma dataset [23], [9], [24], [25], [26]. Fig. 4 shows an improved classification performance for the proposed method in this work (abbreviated as GF&FD).

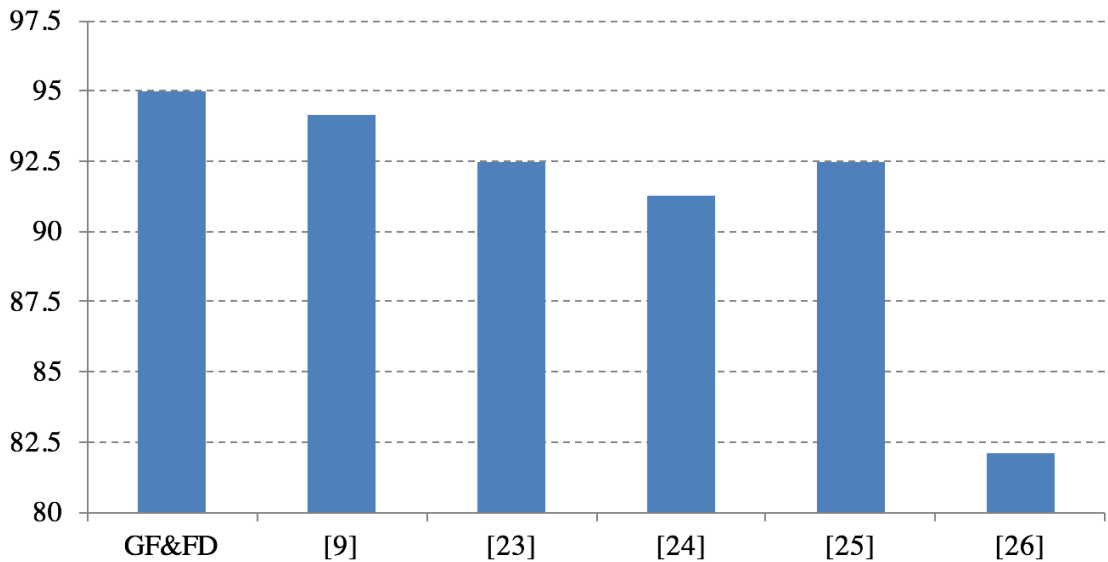

**Fig. 4** Classification performance of the proposed method benchmarked against previous work in [9], [23], [24], [25], [26].

## 4. DISCUSSION

An alternative multiresolution texture analysis approach based on Gabor filters was applied to the meningioma images. Using the energy of the Gabor filter outputs alone as a texture signature for discrimination between the meningioma subtypes proved to be more effective than single or fixed-resolution texture measures. For instance, the highest accuracy achieved for CM, RLM, GMRF, and FD methods used under the same conditions in [23] did not exceed 84% as compared to 89.38% for the Gabor filter in our case. Also, results showed that the Gabor filter classification accuracy could be further improved if combined with the aforementioned texture analysis methods.

The coloured histopathological images were decomposed to the main RGB channels and the blue channel was chosen for operation as dyes used in the staining process of the meningioma tissue biopsies gave the cell nuclei – considered a major mark of distinction between the different subtypes – a colour near to blue. Then feature extraction was performed with and without having the meningioma cell nuclei segmented. The feature extraction involved the multiresolution Gabor filter outputs applied individually for classification and in combination with other fixed-resolution texture measures characterising the same image texture as well. The segmentation process worked well for the filter's output energy signature $G_f(E)_{NS}$ if no segmentation took place. While there was no difference between the segmented and non-segmented versions for the energy and RLM combination $G_f(E\ \&\ RLM)$, it was better to have the histopathological images not segmented when combining the energy of the filter outputs with the FD, GMRF or CM of the image.

All classification results gave an accuracy equal or above 90% when the fixed and multiresolution methods were combined together and without segmentation. The remarkable combination is the $G_f(E\ \&\ FD)_{NS}$ which gave the highest classification accuracy when the subimages were not segmented, see Fig. 4. Although the classification performance degrades when the energy of the $G_f$ was employed without segmentation as indicated in Table II, the loss is compensated and accuracy is further improved when the filter output is combined with the FD of the image. As it was previously shown that the use of the morphological gradient in the segmentation process would negatively affect the classification accuracy of the FD measure [23], it is expected that the fractal characteristics of the non-segmented images to have a better expression which was reflected in the 95.00% accuracy achieved.

It can be deduced that an optimum expression for the meningioma tissue requires both fixed and multiresolution processing to maximize the difference in-between the subtypes. As the cell nuclei size varies between the different subtypes, the multiresolution approach can better identify these differences and represent them by a measurable quantity (e.g. Gabor filter banks energy signature for this work) exemplifying the amount of information at different scales. Yet relying on the energy of the filter outputs alone is deemed insufficient, thus analysing the texture at its highest resolution using a fixed-resolution approach could contribute towards highlighting some of the aspects overlooked by the Gabor filter approach. An example was the FD signature that provides a mean to check for self-similarity or roughness of the surface, where finer texture objects would result in a higher FD estimate and vice versa. Since FD is considered scale invariant [27, 28], it could provide more stable parameter estimation amidst the intra-variability of the number of cell nuclei in the same subtype.

Similarly, the rest of the applied fixed-resolution methods had also its own approach in texture characterisation, whether the emphasis was on the dependence of each pixel in the image only on its neighbours as the case in GMRF, or deriving first and second order statistics after computing the joint probability or the number of runs for a certain set of pixels with a certain grey-level value for CM and RLM; respectively.

Although finer quantisation was recommended by having an orientation angle of 30° for centring the location of the filters [1], a 45° orientation angle was used as its is less computationally expensive and finer orientation quantisation did not show significant improvement for the given histopathological texture samples. Also, experimenting by smoothing the subimages' texture with a Gaussian filter prior to extraction of the feature was applied as well. The approach suggested by Jain and Farrokhnia through setting the filter's window size relative to the radial frequencies of the corresponding tuned filter [16] resulted in an 1% decrease in the recorded accuracy. This can be referred to the blurring which negatively affected the accuracy of the FD signature.

An advantage of the fixed-resolution approach is its simplicity in application and speed in processing. Yet the complexity and non-stationary nature of medical texture requires a multi-perspective analysis in order to spot subtle differences that might be crucial in subtype discrimination. This work takes a step further and demonstrates that a generated fused feature vector exploiting the strengths of both resolution approaches could provide an optimum characterisation for meningioma texture classification. Furthermore, testing the proposed approach on other grades of meningioma or different types of brain tumours would assist in performance benchmarking.

## 5. CONCLUSION

The multiresolution approach based on Gabor filters proved to more effective in terms of classification accuracy as compared to four other fixed-resolution approaches. Additionally, the generation of a feature vector mutually combining the energy signature of the Gabor filter outputs and either of the FD, GMRF, CM or RLM fixed-resolution methods would improve subtype discrimination. The appropriate selection of the feature extraction method(s) according to the nature of the examined tissue texture is necessary for boosting the ability to distinguish subtle differences between the meningioma subtypes, which was optimised by the fusion of the fractal characteristics with the filter bank energy signature. The proposed novel approach could have significant clinical utility by reducing the subjectivity in the process of cancer diagnostics and thereby reducing the pathologist workload and improving diagnosis for patients.